\pdfoutput=1

\documentclass[11pt]{article}

\usepackage[final]{ACL2023}

\usepackage{times}
\usepackage{latexsym}
\usepackage{verbatim}
\usepackage{natbib}
\usepackage{subcaption}  

\usepackage[T1]{fontenc}

\usepackage[utf8]{inputenc}
\usepackage{microtype}
\usepackage{cancel}
\usepackage{inconsolata}

\usepackage{graphicx}
\usepackage[inline]{enumitem}
\usepackage{multirow}

\newcommand{\RS}[1]{\textcolor{violet}{RS: #1}}

\newcommand{\OM}[1]{\textcolor{orange}{OM: #1}}

\title{\textsc{DEplain}: A German Parallel Corpus with Intralingual Translations into Plain Language for Sentence and Document Simplification}

\author{Regina Stodden\textmd{,} Omar Momen \textmd{and} Laura Kallmeyer\\
  Heinrich Heine University \\
  Düsseldorf, Germany \\
  \texttt{\{firstname.secondname\}@hhu.de} \\ 
}

\begin{document}
\maketitle
\begin{abstract}
Text simplification is an intralingual translation task in which documents, 
or sentences of a complex source text are simplified for a target audience. The success of automatic text simplification systems is highly dependent on the quality of parallel data used for training and evaluation. To advance sentence simplification and document simplification in German, this paper presents \textsc{DEplain}, a new dataset of parallel, professionally written and manually aligned simplifications in plain German (\emph{``plain DE'' or in German: `Einfache Sprache''}). \textsc{DEplain} consists of a news-domain (approx. 500 document pairs, approx. 13k sentence pairs) and a web-domain corpus (approx. 150 aligned documents, approx. 2k aligned sentence pairs). In addition, we are building a web harvester and experimenting with automatic alignment methods to facilitate the integration of non-aligned and to be-published parallel
documents. 
Using this approach, we are 
dynamically increasing the web-domain corpus, so it is currently extended to approx. 750 document pairs and approx. 3.5k aligned sentence pairs. We show that using \textsc{DEplain} to train a transformer-based seq2seq text simplification model can achieve promising results. We make available the corpus, the adapted alignment methods for German, the web harvester and the trained models here: \url{https://github.com/rstodden/DEPlain}.

\end{abstract}

\section{Introduction}

Automatic text simplification (TS) is the process of automatically generating a simpler version of complex texts while preserving the main information~\cite{alva-manchego-etal-2020-data}. Current TS research mostly focuses on English and on sentence-level simplification. This paper contributes to TS research on German. Compared to other European languages, German is 
 more difficult to read due to complex sentence structures and 
  many compound words~\cite{marzari-2010-leichtes}. According to 
   \citet{buddeberg-grotluschen-2020-leo}, roughly 6.2 mio. adults in Germany (approx. 12.1\%) have reading and writing problems on the character-level (0.6\%), word-level (approx. 3.4\%) or sentence-level (approx. 8.1\%). 
 To counteract and make texts accessible to more people, currently two dominant German variants for simplified language exist~\cite{maass-2020-easy}:
\begin{enumerate}
    \item \emph{easy-to-read German (de: ``Leichte Sprache'')}: following strict rules  the complexity of the language is maximally reduced (almost corresponds to CEFR level A1). 
    The main target group is people with cognitive or learning disabilities or communication impairments.
    \item \emph{plain German (de: ``Einfache Sprache'')}: reduced complexity with a mild to a strong extent (almost corresponds to CEFR levels A2 and B1), which can be compared to texts for non-experts. The main target group is people with reading problems and non-native German speakers.
\end{enumerate} 
This is also reflected in a rise in research and application of manual and automatic German text simplification:
\begin{enumerate*}[label=\roman*)]
    \item Many German web pages are provided in 
     standard German as well as in 
      plain or easy-to-read German, e.g., Apotheken Umschau\footnote{\href{https://www.apotheken-umschau.de/einfache-sprache/}{https://www.apotheken-umschau.de/einfache-sprache/}} or the German Federal Agency for Food\footnote{\href{https://www.bzfe.de/einfache-sprache/}{https://www.bzfe.de/einfache-sprache/}}, 
    \item News agencies are publishing their news in plain or easy-to-read German, e.g., Austrian Press Agency\footnote{\href{https://science.apa.at/nachrichten-leicht-verstandlich/}{https://science.apa.at/nachrichten-leicht-verstandlich/}} or Deutschlandfunk\footnote{\href{https://www.nachrichtenleicht.de/}{https://www.nachrichtenleicht.de/}}.
\end{enumerate*}

\citet{klaper-etal-2013-building} were the first who made use of these resources for supervised, automatic German TS. 
 They created a small parallel corpus of approx. 250 web pages with intralingual translations from standard German to easy-to-read German. However, due to copyright issues, they (and also its extension by \citet{battisti-etal-2020-corpus}) could not 
  make their corpus publicly available. To avoid such problems, 
  \citet{hewett-stede-2021-automatically,aumiller-gertz-2022-klexikon} built TS 
   corpora based on open accessible Wikipedia texts simplified for children. Due to the high cost of manual sentence-wise alignment or not applicable automatic alignment methods~\cite{aumiller-gertz-2022-klexikon}, these corpora are only aligned on the document level. Furthermore, \citet{spring-etal-2022-ensembling} report results on experiments with some existing automatic alignment methods and show non-satisfying, error-prone results. \par

In this work, we tackle some of the named problems by proposing, \textsc{DEplain}, a new parallel German corpus for text simplification with manual and automatic alignments on the document and sentence level. \textsc{DEplain} contains intralingual translations mostly into plain German and includes ``strong'' as well as ``mild'' simplifications. Overall, we propose 4 subcorpora with in total 1,239 document pairs,  14,968 manual sentence-wise alignments and 1,594 automatic sentence-wise alignments. One subcorpus is built from professionally simplified news articles in plain language of the Austrian Press Agency\footnote{\url{https://science.apa.at/nachrichten-leicht-verstandlich/}}. The resources of the other 3 subcorpora are compiled by a new web harvester, making use of publicly available parallel documents. We analyze these subcorpora based on human ratings and annotations to get more insights into the quality and the simplification processes within the data. 

We further show two use cases of our new TS corpus: 
\begin{enumerate*}[label=\roman*)]
    \item evaluating automatic alignment methods, and
    \item exemplifying TS training and evaluation with \textsc{DEplain}. 
\end{enumerate*}
Our data, web harvester, code for alignment methods and models are publicly available (with some restrictions).

\section{Related Works}
\paragraph{Text Simplification (TS)} is an NLG task in which mostly machine learning models learn from 
complex-simple pairs how to simplify texts for a specific target group. For a lot of languages, 
parallel TS corpora exist either on sentence-level or document-level.\footnote{For an extensive overview see \citet{stajner-2021-automatic-repaired} or \citet{Trienes-etal-2023-datasets}.} Only a few corpora contain data on both levels, e.g., EW-SEW v2.0~\cite{kauchak-2013-improving}, Newsela 2015~\cite{xu-etal-2015-problems}, or Wiki-Auto~\cite{jiang-etal-2020-neural}. Newsela~\cite{xu-etal-2015-problems}, furthermore, includes for each source text several simplified versions targeted to different audiences. Therefore, it contains ``strong'' simplifications (highest to lowest complexity level) and also ``mild'' simplifications (intermediate complexity levels)~\cite{stajner-etal-2017-sentence}. In this paper, we also introduce one corpus with rather ``mild'' simplifications (\textsc{DEplain-apa}) and one with rather ``strong'' simplifications (\textsc{DEplain-web}) which allows more analysis of the capabilities of TS models.
Like most other existing TS corpora (see \citealt{Trienes-etal-2023-datasets}), our corpus contains only one golden simplification (reference) per simplification pair.

\paragraph{German corpora} were also proposed in recent years, both on either document level (e.g., Lexika-corpus~\citealt{hewett-stede-2021-automatically}, or 20Minuten~\citealt{rios-etal-2021-new}) or sentence-level (e.g., web-corpus, APA-LHA, capito-Corpus~\citealt{ebling-etal-2022-automatic}, or Simple-German-Corpus~\citealt{toborek-etal-2022-new}), but not focussing on both levels\footnote{For an extensive overview of existing German datasets including meta-data see \autoref{table-german-datasets}.}. Unfortunately, many of the datasets cannot be used for training TS models or only with caution because, for example, 
\begin{enumerate*}[label=\roman*)]
    \item they are too small for training (e.g., \citet{klaper-etal-2013-building}), 
    \item they are automatically aligned with questionable quality (e.g., \citealt{spring-etal-2021-exploring}), 
    \item are only available for evaluation (e.g., \citealt{mallinson-etal-2020-zero}; \citealt{naderi-etal-2019-subjective}),
    \item they are not truly parallel as the complex and simple versions are written independently (e.g., \citealt{aumiller-gertz-2022-klexikon}), or
    \item are not available (e.g.,~\citealt{ebling-etal-2022-automatic}) sometimes due to copyright issues (e.g.,~\citealt{battisti-etal-2020-corpus}). 
\end{enumerate*}
\textsc{DEplain} tackles  all of the mentioned problems, i.e., size, alignment quality, simplification quality, and availability.

\paragraph{Alignment Methods} and web scraping are already used to overcome some of these issues. For example, similar to our work, \citet{toborek-etal-2022-new} present a web scraper to scrape parallel  documents from the web and automatically align them. However, \cite{spring-etal-2022-ensembling} have shown that automatic sentence alignment is still an open challenge for German by comparing some existing alignment methods.
In this work, we will evaluate the alignment methods on our manually aligned data and will adapt them for our purpose, e.g., use German resources and incorporate $n$:$m$ alignments. \par

\begin{table*}[htb]\resizebox{0.95\textwidth}{!}{

\begin{tabular}{ll|lll|ll}
\textbf{Name} & \textbf{License} & \textbf{\# Doc. Pairs} & \textbf{\# Original Sents} & \textbf{\# Simple Sents.} & \textbf{Alignment} & \textbf{\# Sent. Pairs} \\ \hline\hline
\textbf{\textsc{DEplain-apa}} & upon request & 483 & 25,607 & 26,471 & manual & 13,122 \\ \hline
\multirow{3}{*}{\textbf{DEplain-web}} & open & 147 & 6,138 & 6,402 & manual & 1,846 \\
 & open & 249 & 7,087 & 7,760 & auto & 652 \\
 & closed & 360 & 12,847 & 18,068 & auto & 942 \\ \hline
\textbf{In total} & mixed & 1,239 & 51,681 & 58,701 & mixed & 16,562
\end{tabular}
}
\caption{Overview of the corpora of \textsc{DEplain} including meta data. 
} 

\label{deplain-corpora}
\end{table*}

\section{Document-level TS Corpora}
We present two new TS corpora on the document level, \textsc{DEplain-apa} 
 and \textsc{DEplain-web}, 
  containing 
  parallel documents in standard German and plain German. 
  \autoref{deplain-corpora} provides statistics of both corpora.

\subsection{\textsc{DEplain-apa}}
\label{sec-deplain-apa}
The Austrian Press Agency (APA) publishes everyday news in standard German and parallel, professionally simplified versions for German language learners of CEFR level B1 and A2 (both equivalent to plain language):

\textsc{DEplain-apa} contains news text of APA of CEFR level A2 and B1 which were published between May 2019 and April 2021. Data from the same source was already used for experiments with TS (see for an overview \citet{ebling-etal-2022-automatic}) and made available as APA-LHA~\cite{spring-etal-2021-exploring}. \\
However, the APA-LHA alignments  have some issues that are problematic for training a TS system: The alignment format is unclear in the sense of not distinguishing between $1$:$1$, $1$:$m$, and $n$:$1$ sentence alignments. Furthermore, the documents were aligned automatically, which results in many misaligned sentence-level alignments. Some examples of these problems are presented in \autoref{appendix-apa-lha-errors}.\par

We tackle these problems by making use of the provided manual document alignments of APA from Common European Framework of Reference for Languages (CEFR) level B1 to A2. As the document alignments are not available for all APA documents, our corpus is reduced to 483 parallel documents. In a further comparison, \textsc{DEplain-apa} focuses more on mild simplifications which might be easier to learn for a document TS system than strong simplifications as in APA-LHA (C2 to B1 and C2 to A2).\footnote{Examples for strong and mild simplifications of \textsc{DEplain} can be found in \autoref{appendix-mild-strong-examples}.}

Overall, \textsc{DEplain-apa} contains 483 document pairs (see \autoref{appendix-overview-corpora} and \autoref{table-german-datasets-deplain}). On average, the complex documents (CEFR-level B1) have a German Flesch-Reading-Ease score (FRE)~\cite{flesch-1948-new,amstad-1978-verstandlich} of 61.05 $\pm$ 4.67 and simple documents of 66.48 $\pm$ 4.56, which can be both interpreted as \textit{simple} (following \citet[p. 117]{amstad-1978-verstandlich}).\footnote{We calculated the German variant of FRE with the Python package textstat.  For criticism on traditional readability scores for TS see, e.g., \citet{tanprasert-kauchak-2021-flesch}.}

\subsection{\textsc{DEplain-web}}
\label{sec-deplain-web}
The second document-level corpus of DEplain, i.e., \textsc{DEplain-web}, is a dynamic corpus with 
parallel documents in standard German and plain German from the web. Similar to \citet{battisti-etal-2020-corpus} and \citet{toborek-etal-2022-new}, we have built an open-source web harvester in Python to download, align and extract text of 
parallel documents of given web pages (including paragraphs). For reproducibility, we made the code and the list of web pages available.
However, the automatic extraction of the web data is not perfect as some recent changes in the HTML structure are not recognized by the crawler, and some layouts such as tables or lists might not be extracted correctly. Following this, the data might include some low-quality data.\par

\textsc{DEplain-web} currently contains 756 parallel documents 
 crawled from 11 web pages and covering 
  6 different domains: fictional texts (literature and fairy tales), bible texts, health-related texts, texts for language learners, texts for accessibility, and public authority texts. The first three domains are not 
 included in any other German TS corpus. \par
All simplified documents are professionally simplified by trained translators and often 
  reviewed by the target group.  The simplified documents of 5 of the 11 web pages are written in plain German, 6 in 
   easy-to-read German. All complex documents 
    are in standard German, except \emph{Alumniportal Deutschland}, which contains data on CEFR level B2. 
    Some of the fictional complex documents are only available in their original language from the 19th century and are therefore 
     more difficult to read. 
 More details on the scraped web pages are given  in 
  \autoref{sec-deplain-web-process}. We plan further extensions of \textsc{DEplain-web}, 
   e.g., by a political lexicon in plain German\footnote{\href{https://www.bpb.de/kurz-knapp/lexika/lexikon-in-einfacher-sprache/}{https://www.bpb.de/kurz-knapp/lexika/lexikon-in-einfacher-sprache/}}. \par
   The corpus is dynamic for three reasons: 
  \begin{enumerate*}[label=\roman*)]
      \item it can be extended with new web pages, 
      \item the number of parallel documents of a web page can change, and
      \item the content of the considered web pages can change over time.
  \end{enumerate*} 
    More details on the web crawler, reasoning for choosing the current web pages, and the document alignment process can be found in \autoref{sec-deplain-web-process}.\par

On the one hand, some of these web documents are openly licensed and some data providers allowed us to use and share the data for academic purposes. Therefore, we can publicly share this data; this corpus contains 396 document pairs which are represented in the second and third rows in \autoref{deplain-corpora}. On the other hand, we additionally provide the web crawler to download and use the parallel 
documents with restricted licenses (360 documents) which is represented in the last row in \autoref{deplain-corpora}. \par

\section{Sentence-level TS Corpora}
We aligned both corpora, \textsc{DEplain-apa} and \textsc{DEplain-web}, also on the sentence level. All 483 available parallel documents of \textsc{DEplain-apa}  and 147 documents of \textsc{DEplain-web} are manually aligned on the sentence level with the assistance of a TS annotation tool. Overall, 14,968 sentence pairs of 630  document pairs are manually aligned. We first describe the annotation procedure (see \autoref{sec-annotation-process}) and the resulting statistics per subcorpus (\autoref{sec-deplain-apa-sent} and \autoref{sec-deplain-web-sent}).

\subsection{Annotation Procedure}
\label{sec-annotation-process}
\textsc{DEplain-apa} and \textsc{DEplain-web} are both annotated following the same procedure. The sentence pairs are manually aligned by two German native speakers\footnote{Both annotators  were paid for their work with at least the minimum wage of their country of residence.} using the TS annotation tool TS-anno~\cite{stodden-kallmeyer-2022-ts} which assist, for example, in splitting the documents into sentences, alignment of $n$:$m$ sentence pairs, automatic alignment of identical sentence pairs, and the annotation of simplification operations and manual evaluation. The annotators were instructed by the principal investigator and were also provided with instructions on how to use the annotation tool and with an annotation guideline.\footnote{We adapted the annotation schema of \citet{stodden-kallmeyer-2022-ts} to our needs.} 
\paragraph{Sentence-wise Alignment}
 The manual sentence-wise alignments reflect all possible alignment types: 
\begin{enumerate*}[label=\roman*)]
    \item $1$:$1$ (rephrase and copy), 
    \item $1$:$m$ (split of a complex sentence), 
    \item $n$:$1$ (merge of complex sentences), 
    \item $n$:$m$ (where $n$ and $m>1$, fusion of complex and simple sentences).
\end{enumerate*}
Furthermore, all not annotated sentences of an annotated document are either treated as 
\begin{enumerate*}[label=\roman*)]
  \setcounter{enumi}{4}
    \item $1$:$0$ (deletion of a complex sentence), and 
    \item $0$:$1$ (addition of a simplified sentence).
\end{enumerate*}\par
In the alignments of \textsc{DEplain-apa} and \textsc{DEplain-web}, the complex documents are fully aligned with the simplified documents. This means the alignments also reflect deletions  and additions.

The publication of the full document alignments, also enhance the option for example, 
\begin{enumerate*}[label=\roman*)]
    \item to build a simplification plan for document-level simplification using sequence labeling (see \citealt{cripwell-etal-2023-document}), 
    \item to include preceding and following sentences for context-aware sentences simplification (see \citealt{sun-etal-2020-helpfulness}), or 
    \item to use identical pairs and additions as augmented data during training (see \citealt{palmero-aprosio-etal-2019-neural}).
\end{enumerate*}

\paragraph{Agreement of Alignment and Data Cleaning}
To compare the agreement of both annotators, we randomly sampled 99 documents over all domains which were annotated by both annotators. For calculating the inter-annotator-agreement we framed the alignment as a classification task in which a label (not aligned, partially aligned, or aligned) is assigned to each combination of complex sentences and all simple sentences per document. This format was proposed by \citet{jiang-etal-2020-neural}  for training and evaluating a sentence-wise alignment algorithm. The  inter-annotator-agreement (measured with Cohen's $\kappa$) for these documents is equal to approx. 0.85 (n=87645 sentence combinations) which corresponds to a strong level of agreement (following \citet[p. 279]{mchugh2012interrater}). The lowest agreement is shown for the domain of health data ($\kappa$=0.52, n=13736, interpretation: weak) whereas the highest agreement is shown for the language learner data ($\kappa$=0.91, n=18493, interpretation: almost perfect).\footnote{In \autoref{appendix-tab-iaa}, we show all inter-annotator agreements per domain. } 
The health data was strongly and independently written in plain language (not sentence by sentence), including moving sentences from document beginning to ending or sentence fusion. Therefore, the manual alignment of strong simplifications seems to be less congruent than for the very mild simplifications of the language learner data which have a low edit distance. 

As the texts of \textsc{DEplain-web} are automatically extracted from the websites, and the documents of both, \textsc{DEplain-apa} and \textsc{DEplain-web}, were automatically split into sentences, the sentence pairs can contain some sentences that are wrongly split. Therefore, we cleaned the dataset and removed too short sentences (e.g., ``Anti-Semitismus.'', engl.: ``Antisemitism.'') and too similar sentences with only one character changed (e.g., complex: ``Das ist schön\underline{!}'', simple: ``Das ist schön\underline{.}'', engl.: ``That is nice.''). Furthermore, some sentence pairs (especially term explanations in the news dataset, n=1398) are repeated several times in different documents, we decide to remove all duplicates  to make sure that only unseen sentence pairs occur in the test data set.

\paragraph{Linguistic Annotation}
After cleaning the data, similar to \cite{cardon-etal-2022-linguistic}, 
some randomly selected sentence pairs are annotated with additional linguistic annotations to get more insights into the simplification process of the aligned sentence pairs. We follow the annotation guideline provided in \citet{stodden-kallmeyer-2022-ts}\footnote{The full annotation guideline can be found here: \url{https://github.com/rstodden/TS_annotation_tool/tree/master/annotation_schema}}. We built a typology on linguistic-based operations, which are performed during the simplification process, following a literature review of existing typologies \citet{bott-saggion-2014-text,brunato-etal-2015-design,gonzalez-dios-etal-2018-corpus,koptient-etal-2019-simplification}. Our typology includes 8 operations, i.e., 
\begin{enumerate*}[label=\roman*)]
    \item delete,
    \item insert,
    \item merge,
    \item reorder,
    \item split,
    \item lexical substitution,
    \item verbal changes, and
    \item no changes
\end{enumerate*}
of which each can be annotated on the paragraph-level, sentence-level, clause-level, or word-level. Furthermore, we also manually evaluated the sentence-wise pairs on a few aspects. As no standards for manual evaluation exist~\cite{alva-manchego-etal-2020-data}, we decided to evaluate on the following three most often used criteria, 
\begin{enumerate*} [label=\roman*)]
    \item grammaticality$^\dagger$, 
    \item meaning preservation$^\ddagger$, and  
    \item overall simplicity$^\ddagger$,
\end{enumerate*} and the following additional aspects:
\begin{enumerate*}[label=\roman*)]
    \setcounter{enumi}{3}
    \item coherence$^\dagger$, 
    \item lexical simplicity$^\ddagger$,
    \item structural simplicity$^\ddagger$ (similar to \citet{sulem-etal-2018-semantic}), and
    \item readability (or simplicity)$^\dagger$ (similar to \citet{brunato-etal-2018-sentence}).
\end{enumerate*} 
All aspects marked with $^\ddagger$ are rated on the sentence pair whereas all aspects with $^\dagger$ are rated on the complex as well as the simplified part of the sentence pair. These aspects are rated on a 5-point Likert-scale, to be more clear in the meaning of the scale, the scale either range from -2 to +2 or 1 to 5 following \citet{stodden-2021-scale-unclear}.
Following \citet{alva-manchego-etal-2020-asset,maddela-etal-2021-controllable}, we provide a statement per aspect on which the annotators are asked to agree or disagree on. An overview of the statement per aspect is added to \autoref{appendix-rating_en}.

\subsection{\textsc{DEplain-apa}}
\label{sec-deplain-apa-sent}
\paragraph{Alignment Statistics}
For the sentence-level part of \textsc{DEplain-apa}, all 483 parallel documents are manually aligned following the annotation procedure described above. Overall the subcorpus contains 13,122 manually aligned sentence pairs with 14,071 complex aligned sentences (55.82\% of all complex sentences), and 16,505 simple aligned sentences (63.38\% of all simple sentences) (see \autoref{deplain-corpora}, and \autoref{table-german-datasets-deplain}).

The largest part of the aligned sentence pairs are rephrasings, 75.54\% of the pairs are $1$:$1$ aligned (excluding identical pairs). 17.99\% of the complex sentences are split into several simpler sentences ($1$:$m$ alignments) and 2.91\% were merged into one simple sentence. The remaining 3.57\% are a fusion of several complex and several simple sentences (see \autoref{appendix-deplain-statistics}). Overall, the average sentence length has increased during simplification (complex: 12.64, simple: 13.02) which might be due to splitting long compound words into several tokens.

\paragraph{Manual Evaluation.}
For manual evaluation of \textsc{DEplain-apa}, 46 randomly sampled sentences were rated. The ratings (see \autoref{table-rating-full}) confirm that the corpus contains rather mild simplifications: the original sentences are already simple (4.39$\pm0.77$, max=5) and they are only simplified a bit (0.57$\pm0.86$). Furthermore, the original and the simplified sentences are very grammatical (complex:1.96$\pm$0.29, simple: 2.0$\pm$0.0), rather coherent (complex:3.26$\pm$1.6, simple: 3.54$\pm$1.54), and preserve the meaning (4.33$\pm$0.97).

\begin{table*}[htb]
\resizebox{\textwidth}{!}{
\begin{tabular}{ll|l|l|l|l||ll|ll|ll}
\textbf{} &  & \textbf{Simplicity} & \textbf{LexSimp} & \textbf{StructSimp} & \textbf{MeaningP.} & \multicolumn{2}{c|}{\textbf{Coherence}}  & \multicolumn{2}{c|}{\textbf{Grammaticality}} & \multicolumn{2}{c}{\textbf{Simplicity}} \\
 &  & \textbf{sent. pair} & \textbf{sent. pair} & \textbf{sent. pair} & \textbf{sent. pair} & \textbf{complex} & \textbf{simple} & \textbf{complex} & \textbf{simple} & \textbf{complex} & \textbf{simple} \\
\textbf{corpus} & \textbf{n} & \textbf{(-2 to +2)} & \textbf{(-2 to +2)} & \textbf{(-2 to +2)} & \textbf{(1 to 5)} & \textbf{(1 to 5)} & \textbf{(1 to 5)} & \textbf{(-2 to +2)} & \textbf{(-2 to +2)} & \textbf{(1 to 5)} & \textbf{(1 to 5)} \\ \hline\hline
\textbf{\textsc{apa}} & 46 & 0.57$\pm$0.86 & 0.28$\pm$0.54 & 0.5$\pm$0.81 & 4.33$\pm$0.97 & 3.26$\pm$1.6 & 3.54$\pm$1.54 & 1.96$\pm$0.29 & 2.0$\pm$0.0 & 4.39$\pm$0.77 & 4.72$\pm$0.46 \\ \hline
\textbf{\textsc{web}} & 384 & 1.04$\pm$0.82 & 0.67$\pm$0.75 & 0.95$\pm$0.87 & 4.29$\pm$0.93 & 2.82$\pm$1.48 & 3.08$\pm$1.4 & 1.72$\pm$0.79 & 1.96$\pm$0.26 & 3.48$\pm$1.18 & 4.46$\pm$0.69 \\
\textbf{news} & 46 & 0.57$\pm$0.86 & 0.28$\pm$0.54 & 0.5$\pm$0.81 & 4.33$\pm$0.97 & 3.26$\pm$1.6 & 3.54$\pm$1.54 & 1.96$\pm$0.29 & 2.0$\pm$0.0 & 4.39$\pm$0.77 & 4.72$\pm$0.46 \\ \hline
\textbf{bible} & 155 & 1.39$\pm$0.68 & 0.98$\pm$0.78 & 1.28$\pm$0.77 & 4.34$\pm$0.84 & 2.12$\pm$1.22 & 2.63$\pm$1.22 & 1.45$\pm$1.06 & 1.92$\pm$0.35 & 2.97$\pm$1.27 & 4.44$\pm$0.72 \\
\textbf{lang.} & 157 & 0.67$\pm$0.74 & 0.36$\pm$0.57 & 0.57$\pm$0.73 & 4.46$\pm$0.73 & 3.83$\pm$1.27 & 3.82$\pm$1.27 & 1.96$\pm$0.22 & 1.97$\pm$0.21 & 4.01$\pm$0.81 & 4.43$\pm$0.71 \\
\textbf{fiction} & 72 & 1.1$\pm$0.95 & 0.69$\pm$0.78 & 1.08$\pm$1.02 & 3.82$\pm$1.29 & 2.08$\pm$1.06 & 2.42$\pm$1.33 & 1.75$\pm$0.71 & 2.0$\pm$0.0 & 3.42$\pm$1.16 & 4.56$\pm$0.58
\end{tabular}}
\caption{Results (mean and standard deviation) of the manual evaluation of the manually aligned sentence pairs per subcorpus (upper) and domain (lower). The left part contains results of aspects on the sentence pair (simplicity, lexical simplicity (LexSimp), structural simplicity (StructSimp), and meaning preservation (MeaningP.)) and the right part for the original and simplified sentences (coherence, grammaticality, and simplicity).
}
\label{table-rating-full}
\end{table*}

\paragraph{Simplification Operations.}
184 sentence pairs were annotated with their transformations, for some sentence pairs more than one transformation was performed at the same time. 47.83\% of the pairs are changed on the sentence level and in 84.24\% a change was performed on the word level. On the sentence level, most often a sentence was reordered (48.86\%), split (35.23\%), or rephrased (12.5\%). On the word level, most often a lexical substitution was performed (84.24\%), a word added (46.45\%) or a word deleted (35.48\%).

\paragraph{Interpretation \& Summary}
This analysis shows that the simplifications of \textsc{DEplain-apa} are of a high quality (grammaticality, meaning preservation, coherence) and that they contain a lot of different simplification strategies (e.g., reordering, splitting, lexical substitution). So even if they are labeled as ``mild'' simplifications due to their close language levels (B1 to A2), they seem to be very valuable for training a TS corpus.

\subsection{\textsc{DEplain-web}}
\label{sec-deplain-web-sent}
For the sentence-level part of \textsc{DEplain-web}, 147 of the 456 parallel documents are manually sentence-wise aligned. The manual alignment process resulted in 1,846 sentence pairs 
(see \autoref{deplain-corpora} and \autoref{appendix-overview-corpora}).

\paragraph{Alignment Statistics.}
In contrast to \textsc{DEplain-apa}, both the complex sentences (avg$_{web}$=22.59, avg$_{APA}$=12.64) and the simplified sentences (avg$_{web}$=19.76, avg$_{APA}$=13.02) are longer on average, which is due to the different complexity levels (in web, complex is comparable to CEFR level C2, and A2-B2 for simple documents). Following that, the sentence pairs of \textsc{DEplain-web} are more often split (43.12\%) than \textsc{DEplain-apa} (17.99\%). However, still the most often alignment type is the $1$:$1$ alignment (46.86\%). Only 4.06 \% of the complex sentences are merged and 5.96\% are fused.For more statistics on \textsc{DEplain-web} see \autoref{deplain-corpora},  \autoref{table-german-datasets-deplain} and \autoref{appendix-deplain-statistics}.

\paragraph{Manual Evaluation.}
384 randomly sampled sentence pairs  are rated regarding simplification aspects (see \autoref{table-rating-full}). 
Overall during the simplification process, the sentences were improved in coherence (complex: 2.82$\pm$1.48, simple: 3.08$\pm$1.4), grammaticality (complex: 1.72$\pm$0.79, simple: 1.96$\pm$0.26) and simplicity (complex: 3.48$\pm$1.18,  simple: 4.46$\pm$0.69). 
As presumed before, the original sentences of the bible (2.97$\pm$1.27) and the fictional literature (3.42$\pm$1.16) are more complex than the other original texts (even if not reflected in the FRE)\footnote{This might due to the fact that FRE is build for text level and not sentence level. The calculation seems to fail for some sentences, e.g., ``Anti-Semitismus.'' (engl: ``Antisemitism.'') got a score of -172 (extremely difficult to understand) whereas ``Tom!'' is scored with 120.5 (extremely easy to understand). Therefore, these scores must be interpreted with caution.}. However, their simplicity scores of the simple sentence (bible: 4.44$\pm$0.72, fiction: 4.56$\pm$0.58) are comparable to the scores of the other domains, therefore, these alignments can be seen as ``strong'' simplifications.  Furthermore, they also have a higher average for structural and lexical simplifications than the other domains. Overall, \textsc{DEplain-web} is comparable to \textsc{DEplain-apa} in terms of meaning preservation (\textsc{web}: 4.29$\pm$0.93, \textsc{apa}: 4.33$\pm$0.97) and grammaticality (\textsc{web} simple: 1.96$\pm$0.26, \textsc{apa} simple: 2.0$\pm$0.0), contains stronger simplifications (\textsc{web}: 1.04$\pm0.82$, \textsc{apa}: 0.57$\pm0.86$) but the simplified web texts are less coherent than the simplified news (\textsc{web} simple: 3.08$\pm1.4$, \textsc{apa} simple: 3.54$\pm$1.54).

\paragraph{Simplification Operations.}
350 pairs were rated, 50.57\% on the sentence level and 69.43\% on the word level. The manual annotation corresponds to the automatic calculation of alignment types: the most often change on a sentence level is the split of the sentence (50.57\%). 30.51\% of the pairs are rephrased and 14.69\% are reordered. Interestingly also a high percentage of verbal changes (7.91\%) which include changes from passive to active or subjunctive to indicative. On the word level, again lexical substitution is performed most often (86.42\%), in 24.28\% at least one word is deleted, and in 11.52 \% one word is added.

\paragraph{Interpretation \& Summary}
This analysis again shows a mix of different simplification strategies, including lexical changes as well as syntactical changes. The manual ratings also lead to the assumption that the simplifications are strong and of high quality. Therefore, this corpus can also be a great benefit for German TS. \par

Furthermore, the manually aligned sentence pairs and the document pairs of \textsc{DEplain-web} can be used for evaluating alignment algorithms across different domains. The alignment algorithm can then be used to automatically align the not-aligned documents of \textsc{DEplain-web}. We are showing this process in the next section.

\section{Automatic Sentence-wise Alignment}
To exemplify the usage of the manual alignments and to provide sentence-wise alignments for the unaligned documents of \textsc{DEplain-web} we evaluate different alignment algorithms on the manually aligned data.

\subsection{Alignment Methods}
We evaluated the following alignment methods:
\begin{enumerate*}[label=\roman*)]
    \item \emph{LHA}~\cite{nikolov-hahnloser-2019-large},
    \item \emph{SentenceTransformer}~\cite{reimers-gurevych-2020-making} with \emph{LaBSE}\footnote{\href{https://huggingface.co/sentence-transformers/LaBSE}{https://huggingface.co/sentence-transformers/LaBSE}}~\cite{feng-etal-2022-language}  and \emph{RoBERTa}\footnote{\href{https://huggingface.co/T-Systems-onsite/cross-en-de-roberta-sentence-transformer}{https://huggingface.co/T-Systems-onsite/cross-en-de-roberta-sentence-transformer}}~\cite{conneau-etal-2020-unsupervised}
    \item \emph{VecAlign}~\cite{thompson-koehn-2020-exploiting},
    \item \emph{BertAlign}~\cite{10.1093/llc/fqac089},
    \item \emph{MASSAlign}~\cite{paetzold-etal-2017-massalign}, and
    \item \emph{CATS}~\cite{stajner-etal-2018-cats}.
\end{enumerate*}
Before testing any of these alignment methods, we investigated the implementation of their algorithms and checked for any room for adaptation to benefit our purpose.\footnote{More details on our adaptations can be found in \autoref{appendix-alignment-methods}.}

\subsection{Evaluation of Alignment Methods}
\label{sec-auto-alignment-eval}

We chose 
the subcorpus of \textsc{DEplain-web} that has manual alignments and is open for sharing (second row in \autoref{deplain-corpora}) for evaluating the methods, as it has a sufficient number of alignments representing different domains and different types of alignments ($1$:$1$ and $n$:$m$). The dataset comprises 147 aligned pairs of documents, these complex-simple document pairs were split into 6,138 and 6,402 sentences respectively.
The manual alignment of these sentences resulted in 2,741 alignments, comprising 1,750 $1$:$1$ alignments (out of which are 887 identical pairs), 804 $1$:$m$ alignments, 77 $n$:$1$ alignments, and 110 $n$:$m$ alignments.\footnote{$n$ and $m$ are $>1$ in this context}

For evaluation, we treat the alignment task as a binary classification problem (either aligned or not aligned) and report precision, recall, and F1-score. We do not consider partial alignments within the evaluation. We argue that for curating a finetuning dataset for automatic text simplification systems, having an accurate alignment is more important than missing an accurate one, therefore we value precision over recall. Hence, we also measured the F$_{\beta}$ score with ${\beta=0.5}$ which weighs precision more than recall.

\begin{table}[htb]
\resizebox{\columnwidth}{!}{
\begin{tabular}{l|llll|llll}
 & \multicolumn{4}{c|}{\textbf{$1$:$1$}} & \multicolumn{4}{c}{\textbf{$n$:$m$}} \\
 \hline
\textbf{name} & \textbf{P} & \textbf{R} & \textbf{F$_1$} & \textbf{F$_{0.5}$} & \textbf{P} & \textbf{R} & \textbf{F$_1$} & \textbf{F$_{0.5}$} \\\hline\hline
LHA & .94 & .41 & .57 & .747 & - & - & - & - \\
Sent-LaBSE & \textbf{.961} & .444 & .608 & \textbf{.780} & - & - & - & - \\
Sent-RoBERTa & .960 & .444 & .607 & .779 & - & - & - & - \\

CATS-C3G & .247 & \textbf{.553} & .342 & .278 & - & - & - & - \\ \hline
VecAlign & .271 & .404 & .323 & .290 & .260 & .465 & .333 & .285 \\
BERTalign & .743 & .465 & .572 & .664 & .387 & .561 & .458 & .412 \\
MASSalign & .846 & .477 & \textbf{.610} & .733 & \textbf{.819} & .509 & \textbf{.628} & \textbf{.730}
\end{tabular}}
\caption{Results of the alignment methods with $1$:$1$ (upper part) and $n$:$m$ capabilities (lower part) on sentence-pairs with $1$:$1$ (n=1750, left part) and $n$:$m$ alignments (n=991, right part).}

\label{alignment-methods-evaluation}
\end{table}

\subsection{Results}
\label{sec-deplain-sent-auto}

Three of our studied alignment methods can produce only $1$:$1$ alignments (LHA, SentenceTransformer, CATS), and the other three methods can produce additionally $n$:$m$\footnote{where $n$ or $m$ equals to $1$ in this context, but not both} alignments (VecAlign, BertAlign, MASSAlign). Theoretically, our ideal aligner should be able to produce $n$:$m$ alignments with high precision as splitting and merging are frequent in TS corpora. However, in our experiments, we observed that producing $n$:$m$ alignments is a difficult task. We found that SentenceTransformer using the multi-lingual model LaBSE \cite{feng-etal-2022-language} got very high precise $1$:$1$ results with a fair recall as well (see \autoref{alignment-methods-evaluation}). On the other hand, MASSAlign performed the best on $n$:$m$ results, and also with totally acceptable $1$:$1$ results (see \autoref{alignment-methods-evaluation}). Hence, we concluded that MASSAlign is the most suitable aligner for our use case as it 
\begin{enumerate*}[label=\roman*)]
    \item produces $n$:$m$ alignments and 
    \item has fairly high scores for $1$:$1$ and $n$:$m$ alignments.
\end{enumerate*}
Therefore, we recommend MASSAlign to be used to automatically align the documents which the web crawler can scrape.

\subsection{Corpus Statistics}

Running MASSAlign on our unaligned corpus of \textsc{DEplain-web} results in 1,594 sentence alignments. Following statistics of the manually aligned part of \textsc{DEplain-web} (1,846 aligned pairs of 6,138 complex sentences), theoretically, a perfect aligner should get on average a maximum of 30\% alignments of the complex sentences, which corresponds to 5,980 sentence pairs on the not-aligned documents that we posses (with 19,934 complex sentences). However, as we set our experiments with the aim of getting a precise aligner that values quality over quantity, these expected numbers were much reduced in reality (to approx 8\%).

\section{Automatic Text Simplification}
\label{sec-ats}

To exemplify the usage of \textsc{DEplain} for training and evaluating TS models, we are presenting results on finetuning \emph{long-mBART} on our document-level corpus as well as finetuning \emph{mBART} on our sentence-level corpus, using code provided by \citet{rios-etal-2021-new}\footnote{For our experiments, we use all default parameters as reported on GitHub \url{https://github.com/a-rios/ats-models}. We finetuned all of the TS models only once, and therefore reporting only one result per TS model.}.

\subsection{Data}

We have split the document and sentence pairs of \textsc{DEplain-apa} and \textsc{DEplain-web} into training, development, and testing splits, the sizes of all splits are provided in  \autoref{appendix-train-dev-test}.\footnote{We also provide the data split here: \url{https://github.com/rstodden/DEPlain}.} We are reporting evaluation metrics on the test sets of \textsc{DEplain-apa}, and \textsc{DEplain-web} for both document- and sentence-level systems.More results on other test data sets can be found in \autoref{appendix-tables-results-sentence-level}.

\subsection{Evaluation of Text Simplification}

For evaluation, we use the following automatic metrics provided in the evaluation framework EASSE \cite{alva-manchego-etal-2019-easse}: for simplification, SARI \cite{xu-etal-2015-problems}, for quality and semantic similarity to the target reference, BERTScore Precision (BS-P) is reported \cite{zhang2019bertscore}, and BLEU for meaning preservation \cite{papineni-etal-2002-bleu}, following the recommendations of \citet{alva-manchego-etal-2021-un} regarding TS evaluation on English texts. As our corpus has just one reference and not multiple as English TS corpora, e.g., Newsela~\cite{xu-etal-2015-problems} or ASSET~\cite{alva-manchego-etal-2020-asset}, SARI might not work as good as expected.
Instead of Flesch-Kincaid Grading Level (FKGL) which is built for only English data, we are reporting the German version of Flesch-Reading-Ease (for readability).\footnote{We are using metrics even if it is shown (e.g., for SARI~\cite{alva-manchego-etal-2021-un}, BLEU~\cite{sulem-etal-2018-bleu} and FKGL~\cite{tanprasert-kauchak-2021-flesch}) that they are not perfect for measuring simplification as no other more reliable metrics exist yet. A manual evaluation would be most reliable, but this is out of the scope of this work as we are evaluating the validity of the dataset and not the TS system.}

As baseline we use a src2src-baseline or identity-baseline, which 
As baseline we report the results of a src2src-baseline or identity-baseline in which the complex source sentences are just copied and, hence, the complex sentences are used as original and potential simplification data. We cannot use a reference baseline (tgt2tgt) as used in related works, because \textsc{DEplain} has only one reference to evaluate against, hence, the scores would always result in the highest scores, e.g., 100 for SARI. 

\subsection{Results}
\subsubsection{Document-level Text Simplification System}

The evaluation results of the document simplification systems are summarized in \autoref{table-results-doc}, 
In terms of SARI, all the fine-tuned models are outperforming the Identity baseline src2src on both test sets.\footnote{Further results on another test dataset are added to the \autoref{app:result-doc-level}).}

However, against our hypothesis, the strong simplifications of the web data seems to be easier to be simplified (SARI$>$43) than the mild simplifications of the APA data (SARI$>$35). 

For BLEU score, the higher the score, the more of the content was copied \cite{chatterjee-agarwal-2021-depsym}. So, if the BLEU score is less for the system outputs than for the identity baseline that means that the simplification system has changed something and not only copied. Therefore, it is reasonable that the BLEU score is higher for the src2src baselines than for some system outputs (see \citealt{xu-etal-2016-optimizing}, \citet{sulem-etal-2018-simple}, \cite{chatterjee-agarwal-2021-depsym}).
A human evaluation is required in order to obtain a more reliable assessment of the quality of the results, as although those metrics are the traditionally used metrics in this area, they were originally designed for sentences evaluation, and not documents evaluation.

\begin{table}
\begin{subtable}[c]{0.5\textwidth}
\resizebox{\columnwidth}{!}{

\begin{tabular}{l|llllll}
\textbf{train data} & \textbf{n} & \textbf{SARI $\uparrow$} & \textbf{BLEU $\uparrow$} & \textbf{BS-P $\uparrow$} & \textbf{FRE $\uparrow$} \\\hline\hline
DEplain-APA & 387 &  \textbf{44.56} & \textbf{38.136} &  \textbf{0.598} & \textbf{65.4} \\
DEplain-web & 481 &  35.02 & 12.913 & 0.475 & 59.55 \\
DEplain-APA+web & 868 & 42.862 & 36.449 & 0.589 & 65.4 \\\hline
src2Src-baseline &  & 17.637 & 34.247 & 0.583 & 58.85
\end{tabular}}
\subcaption{\textsc{DEplain-apa} test (n=48)}
\end{subtable}
\newline
\vspace*{0.25 cm}
\newline
\begin{subtable}[c]{0.5\textwidth}
\resizebox{\columnwidth}{!}{
\begin{tabular}{l|llllll}
\textbf{train data} &\textbf{n}& \textbf{SARI $\uparrow$} & \textbf{BLEU $\uparrow$} & \textbf{BS-P $\uparrow$} & \textbf{FRE $\uparrow$} \\\hline\hline
DEplain-APA & 387 & 43.087 & 21.9 & 0.377 & \textbf{64.7} \\
DEplain-web & 481 & 49.584 & 23.282 & \textbf{0.462} & 63.5 \\
DEplain-APA+web & 868 & \textbf{49.745} & \textbf{23.37} & 0.445 & 57.95 \\\hline
src2Src-baseline &  & 12.848 & 23.132 & 0.432 & 59.4
\end{tabular}}
\subcaption{\textsc{DEplain-web} test (n=147)}
\end{subtable}
\caption{Results on Document Simplification using finetuned long-mBART. n corresponds to the length of the training data. 
}
\label{table-results-doc}
\end{table}

\subsubsection{Sentence-Level Text Simplification System}
The evaluation results of the sentence level systems are summarized in \autoref{table-results-sent}.\footnote{Further results on other test data are added to  \autoref{appendix-tables-results-sentence-level}.} 
These results can be seen as baselines for further experiments with the \textsc{DEplain} corpus. \par
Comparing the FRE of \textsc{DEplain-apa} and \textsc{DEplain-apa+web} on the two test sets, the \textsc{DEplain-apa} test always achieves a higher FRE. 
Also the model that was trained on APA+web data did not make a big difference from the one trained only on APA when tested on the APA test set; the additional web data does not affect the model much in this case. However, when tested on the web test set, adding the web data to the training data has improved all the measured metrics. This supports that adding training data from different contexts leads to a better generalization of the model.
The combination of \textsc{DEplain-apa+web} achieves the highest scores in terms of all metrics. \par

Although our data doesn't include simplifications for children, the SARI scores on the ZEST-test set are better than the reported models (increase by approx.  5 points on SARI, see appendix~\ref{app-result-sent-level}, \autoref{appendix-table-results-zest}). This result might be due to issues with the automatic TS metrics. A manual evaluation is required to justify if finetuning mBART on \textsc{DEplain-apa+web} can really simplify texts for children. The target group of the texts a model is trained on should always be considered in the interpretation of model evaluation as each target group requires different simplification operations \cite{gooding-2022-ethical}.

\begin{table}
\begin{subtable}[c]{0.5\textwidth}
\resizebox{\columnwidth}{!}{
\begin{tabular}{l|llllll}
\textbf{train data} & \textbf{n} & \textbf{SARI $\uparrow$} & \textbf{BLEU $\uparrow$} & \textbf{BS-P $\uparrow$} & \textbf{FRE $\uparrow$} \\\hline\hline
DEplain-APA & 10660 & 34.818 & 28.25 & 0.639 & \textbf{63.072} \\
DEplain-APA+web & 11941 & \textbf{34.904} & \textbf{28.506} & \textbf{0.64} & 62.669 \\\hline
src2src-baseline &  & 15.249 & 26.893 & 0.627 & 59.23
\end{tabular}}
\subcaption{\textsc{DEplain-apa} test (n=1231)}
\end{subtable}
\newline
\vspace*{0.25 cm}
\newline
\begin{subtable}[c]{0.5\textwidth}
\resizebox{\columnwidth}{!}{
\begin{tabular}{l|llllll}
\textbf{train data }& \textbf{n} & \textbf{SARI $\uparrow$} & \textbf{BLEU $\uparrow$} & \textbf{BS-P $\uparrow$} & \textbf{FRE $\uparrow$} \\\hline\hline
DEplain-APA & 10660 & 30.867 & 15.727 & 0.413 & 64.516 \\
DEplain-APA+web & 11941 & \textbf{34.828} & \textbf{17.88} & \textbf{0.436} & \textbf{65.249} \\\hline
src2src-baseline &  & 11.931 & 20.85 & 0.423 & 60.825
\end{tabular}}
\subcaption{\textsc{DEplain-web} test (n=1846)}
\end{subtable}
\caption{Results on Sentence Simplification using finetuned mBART. n corresponds to the length of the training data. 
}

\label{table-results-sent}
\end{table}

\section{Conclusion and Future Work}
In this paper, we have introduced a new German corpus for text simplification, 
called DEplain. The corpus contains data for document simplification as well as sentence simplification of news (\textsc{DEplain-apa}) and web data (\textsc{DEplain-web}). 
The major part of the sentence-wise alignments are manually aligned and a part of it is also manually analyzed. The analysis shows that the subcorpus \textsc{DEplain-apa} contains rather mild simplifications whereas \textsc{DEplain-web} contains rather strong simplifications. However, for both corpora, a large variety of simplification operations were identified. \\
Furthermore, we evaluated automatic sentence alignment methods on our manually aligned data. In our experiments, MASSalign got the best results but it has only identified a few $n$:$m$ alignments (where $n>1$ or $m>1$). One direction for future work is to further investigate $n$:$m$ alignments algorithms for TS corporaand include paragraphs into the automatic alignment process. We also showed first promising experiments on sentence and document simplification. However, these are just simple benchmarks and have only been evaluated with automatic metrics yet. In future work, the results should be verified by manual evaluation and could be improved by 
using more sophisticated approaches.
Finally, we think that \textsc{DEplain} can boost and improve the research in German text simplification, to make more complex texts accessible to people with reading problems.

\newpage

\section*{Acknowledgments}
We gratefully acknowledge  the support of the Austria Presse Agentur (Austrian Press Agency, APA) for providing us with the news documents and Maya Kruse for her work on annotating the corpus. We further thank  NVIDIA for the access to GPUs. This research is part of the PhD-program ``Online Participation'', supported by the North Rhine-Westphalian (German) funding scheme ``Forschungskolleg''.

\section*{Limitations }

However, our work shows some limitations. A major restriction of our data is the different licenses of our proposed dataset, i.e., 
\begin{enumerate*}[label=\roman*)]
\item \textsc{DEplain-apa} can be obtained for free for research purposes upon request,
\item the manual aligned part of \textsc{DEplain-web} and the smaller part of the  automatically aligned sentences are available under open licenses, e.g., CC-BY-4.0,
\item the other automatically aligned sentence pairs and their documents are not allowed to be shared. However, the web harvester and an automatic alignment method can be used to reproduce the document and sentence pairs. 
\end{enumerate*}\par

But, the web crawler does not perform well for all web pages and extracts one-token sentences. While this is not reflected in the manually aligned sentence pairs due to manual quality checks before alignment, this does not happen for the automatic alignment. As a consequential error, the text simplification model would also learn wrong sentence structures and simplifications from this data. Therefore, more time in automatic data cleaning is required. In addition, the web crawler is currently mostly extracting HTML documents and only a few PDF files. The corpus could be increased if existing parallel PDF files would be crawled and correctly extracted. Furthermore, the web crawler just harvest a given set of web pages and do not search in the whole web for parallel German complex-simplified documents such as a general web crawler. Currently, a general web crawler wouldn't add much more parallel data than the proposed one, as this data is scarce at the moment and, if parallel data is available, there is no link between complex and simplified documents and the title of the pages are often that different that they cannot be aligned automatically. However, if in future more parallel texts are available we would like to extend our corpus with a general web crawler to also include more variance within the domains.\par 

Compared to TS corpora in English, e.g., Wiki-Auto or Newsela-Auto \cite{jiang-etal-2020-neural}, \textsc{DEplain} is smaller and is not (fully) balanced in terms of domains. However, we believe that the current size of the dataset is already large enough due to the high proportion of professionally simplified texts and the high-quality of manual alignments. 
Further, we are aiming at increasing the corpus following our dynamic approach, e.g., by extending the capabilities of the web harvester or the alignment algorithms.

Furthermore, the automatic alignment methods currently align mostly $1$:$1$ alignments but $n$:$m$ alignments are important for text simplification and should be considered for training. Therefore, the automatic alignment methods should be improved in the direction of $n$:$m$ alignments. Until then, we would recommend using the automatically aligned sentence pairs only for additional training data.\par
For document simplification, a GPU with at least 24 GB of memory is required to reproduce our results with mBART and at least 16 GB for sentence simplification respectively. However, for other approaches, e.g., unsupervised learning or zero-shot approaches, the experiments with \textsc{DEplain-apa} and \textsc{DEplain-web} on the document- and sentence-level could be performed with less memory. 

For the evaluation of our text simplification models, we are just reporting automatic alignments, although they are mainly built for English TS and their quality is not evaluated on German yet. 
Our datasets has just one reference and not multiple as in ASSET or TurkCorpus, therefore SARI might work not as good as with several references. 
In addition, in some work (e.g., \citet{alva-manchego-etal-2021-un}) it was already shown that the automatic metrics do not perfectly correlate with human judgments, hence, the results of the automatic metrics should be interpreted with caution. It is recommended to manually evaluate the results, but this was out of the scope of this work which mainly focuses on proposing a new dataset and not new TS models.

\newpage
\section*{Ethics \& Impact Statement}
\subsection*{Data Statement}
The data statement for \textsc{DEplain} is available here: \url{https://github.com/rstodden/DEPlain}.

\subsection*{NLP application statement}
\paragraph{Intended use.}
In this work, we propose a new corpus for training and evaluating text simplification models. It is intended to use this corpus for training text simplification models or related works, e.g., text style transfer. The resulting text simplification system is intended to be used to simplify texts for a given target group (depending on the training data). However, the generated simplifications of the TS model might have some errors, therefore they shouldn't be shown to a potentially vulnerable target group before manually verifying their quality and possibly fixing them. The text simplification system could be provided to human translators who might improve and timely reduce their effort in manually simplifying a text.\par
Furthermore, the dataset can be reused for related tasks to TS, this includes, but is not limited to, text leveling, evaluation of alignment methods, evaluation of automatic TS metrics, and analysis of intralingual translations.

\paragraph{Misuse potential \& Failure modes.}
One potential misuse of \textsc{DEplain} is to reverse the input order of the texts into a deep learning model. The resulting system would be able to make the texts even more complex than simplifying them. 
Although a system that would be developed for producing complex texts can be used for beneficial use cases (e.g., generation of more challenging texts for language learners), however, it could be used to obscure pieces of information from some kind of audience on purpose.
Furthermore, the TS system could generate content with low similarity to the complex sentence given as input and, hence, change the meaning of the original text. Due to this, it should always be stated that the simplification is generated automatically and might not reflect the original meaning of the source text.

\paragraph{Biases.}
No biases are known yet.

\paragraph{Collecting data from users.}
When a researcher requests access to the \textsc{DEplain-apa} corpus, the name, the institution, and the email address of the researcher are saved by the authors of the paper. This is required to make the use of the dataset transparent to the data provider, i.e., the Austrian Press Agency.

\paragraph{Environmental Report.}
For the manual alignment and annotation of the corpus, a server with the text simplification tool has run all time during the annotation duration. 
 For the evaluation of all alignment methods, we required less than 1 GPU hour on an NVIDIA RTX A5000 with 24 GB. The experiments with document and sentence simplification, overall, took less than 18 GPU hours on a NVIDIA RTX A5000 with 24 GB.

\bibliography{anthology}
\bibliographystyle{acl_natbib}

\appendix
\newpage

\section*{Appendix}
\label{sec:appendix}

\section{Overview of Existing German TS Corpora}
\label{appendix-overview-corpora}
In \autoref{table-german-datasets}, an overview of existing German text simplification corpora is shown. For \citet{hewett-stede-2021-automatically}  we report the numbers from the updated version of March 2022.

\begin{table*}
\begin{subtable}[c]{\textwidth}
\resizebox{\textwidth}{!}{
  \rowcolors{2}{gray!15}{white}
\begin{tabular}{p{3.5cm}p{3.2cm}|p{4cm}lp{5cm}lllll}
\textbf{Reference} & \textbf{Name} & \textbf{Target Simple} & \textbf{Domain} & \textbf{Availability} & \textbf{\# Docs} & \textbf{\# Sent. Complex} & \textbf{\# Sent. Simple} & \textbf{\# Aligned Pairs} & \textbf{Alignment} \\\hline\hline
\citet{siegel-etal-2019-aspects} & leichte-sprache-corpus  & mixed & web & \url{https://github.com/hdaSprachtechnologie/easy-to-understand_language} & 351 &  &  &  &  \\
\citet{hewett-stede-2021-automatically} & Lexica-corpus-klexikon & children between 6-12 & wikipedia & \url{https://github.com/fhewett/lexica-corpus} & 1090 &  &  &  & auto \\
\citet{hewett-stede-2021-automatically} & Lexica-corpus-miniklexikon & children younger than 6 & wikipedia & \url{https://github.com/fhewett/lexica-corpus} & 1090 &  &  &  & auto \\
\citet{rios-etal-2021-new}$^*$ & 20Minuten & general & news & \url{https://github.com/ZurichNLP/20Minuten} & 18305 &  &  &  & ? \\
\citet{aumiller-gertz-2022-klexikon} & Klexikon & children between 6-12 & wikipedia & \url{https://github.com/dennlinger/klexikon} & 2898 & 701577 & 94214 &  & auto \\
\citet{ebling-etal-2022-automatic} & Wikipedia-Corpus & A2 & wikipedia & - & 106126 & 6933192 & 1077992 &  & ? \\
\citet{trienes-etal-2022-patient} $^\dagger$  & simple-patho & laypeople & medical & \url{https://github.com/jantrienes/simple-patho} (not yet available) & 851 & 23,554 & 28,155 &  \begin{tabular}[c]{@{}l@{}}2,280\\  (paragraphs)\end{tabular}& manual \\ 
\citet{schomacker-2023-aligned} & MILS+EB+PV+KV & mixed & fiction & \url{https://github.com/tschomacker/aligned-narrative-documents} (not yet available) &  &  &  & &  \\ 

\end{tabular}}
\subcaption{Overview of German simplification corpora on document-level. }
\label{table-german-datasets-document}
\end{subtable}
\newline
\vspace*{0.25 cm}
\newline
\begin{subtable}[c]{\textwidth}
\resizebox{\textwidth}{!}{
  \rowcolors{2}{gray!15}{white}
\begin{tabular}{p{3.5cm}p{3.2cm}|p{4cm}lp{5cm}lllll}
\textbf{Reference} & \textbf{Name} & \textbf{Target Simple} & \textbf{Domain} & \textbf{Availability} & \textbf{\# Docs} & \textbf{\# Sent. Complex} & \textbf{\# Sent. Simple} & \textbf{\# Aligned Pairs} & \textbf{Alignment} \\\hline\hline
\citet{klaper-etal-2013-building} & Klaper & Leichte Sprache & web & upon request & 256 &  &  & approx. 2000 & manual\&auto \\
\citet{naderi-etal-2019-subjective} & TextComplexityDE19 & Non-native speaker (written by non-native speaker) & wikipedia & \url{https://github.com/babaknaderi/TextComplexityDE} & 23 &  &  & 250 & manual \\
\citet{battisti-etal-2020-corpus,ebling-etal-2022-automatic} & Web & A2 & web & - & 378 & 17121 & 21072 &  & CATS \\
\citet{mallinson-etal-2020-zero} & ZEST-data & children between 5-7 & \begin{tabular}[c]{@{}l@{}}science for\\  children\end{tabular}  & \url{https://github.com/Jmallins/ZEST-data} & 20 &  &  & 1198 & manual \\
\citet{sauberli-etal-2020-benchmarking,ebling-etal-2022-automatic} $^\ddagger$ & APA-benchmark & B1 & news & - &  &  &  & 3616 & CATS-WAVG \\
\citet{kim-etal-2021-bisect}  & BiSECT &  & \begin{tabular}[c]{@{}l@{}}web \&\\ politics \end{tabular}& \url{https://github.com/mounicam/BiSECT}  &  &  &  & 186,237 &  \\
\citet{hansen-schirra-etal-2021-intralingual} & GEASY & Leichte Sprache & mixed & - & 93 & 1596 & 4090 &  & \begin{tabular}[c]{@{}l@{}}memsource \&\\ (commercial) \end{tabular} \\
\citet{spring-etal-2021-exploring,ebling-etal-2022-automatic} $^\ddagger$ & APA-LHA-or-a2 & A2 & news & \url{https://zenodo.org/record/5148163} & 2426 & 60732 & 30432 & 9456 & LHA \\
\citet{spring-etal-2021-exploring,ebling-etal-2022-automatic} $^\ddagger$ & APA-LHA-or-b1 & B1 & news & \url{https://zenodo.org/record/5148163} & 2426 & 60732 & 30328 & 10268 & LHA \\
\citet{spring-etal-2021-exploring,ebling-etal-2022-automatic} & capito & B1 & news & - & 1055 & 183216 & 68529 & 54224 & LHA \\
\citet{spring-etal-2021-exploring,ebling-etal-2022-automatic} & capito & A2 & news & - & 1546 & 183216 & 168950 & 136582 & LHA \\
\citet{spring-etal-2021-exploring,ebling-etal-2022-automatic} & capito & A1 & news & - & 839 & 183216 & 24243 & 10952 & LHA \\
\citet{toborek-etal-2022-new} & Simple German Corpus & A1 & web & \url{https://github.com/buschmo/Simple-German-Corpus} & 530 &  &  & 5889 & CATS

\end{tabular}}
\subcaption{Overview of German simplification corpora on sentence-level. }
\label{table-german-datasets-sentence}
\end{subtable}
\newline
\vspace*{0.25 cm}
\newline
\begin{subtable}[c]{\textwidth}
\resizebox{\textwidth}{!}{
  \rowcolors{2}{gray!15}{white}
\begin{tabular}{p{3.5cm}p{3.2cm}|p{4cm}lp{5cm}lllll}
\textbf{Reference} & \textbf{Name} & \textbf{Target Simple} & \textbf{Domain} & \textbf{Availability} & \textbf{\# Docs} & \textbf{\# Sent. Complex} & \textbf{\# Sent. Simple} & \textbf{\# Aligned Pairs} & \textbf{Alignment} \\\hline\hline
 & \textsc{DEplain-apa} $^\ddagger$ & A2 & news & \url{https://github.com/rstodden/DEPlain} & 483 & 14,071 (aligned) & 16,505 (aligned) & 13122 & manual\\
 & \textsc{DEplain-web} & mixed & web & \url{https://github.com/rstodden/DEPlain} & \begin{tabular}[c]{@{}l@{}}147\\ (+609)\end{tabular} & 2,287 (aligned)& 4,009 (aligned)& \begin{tabular}[c]{@{}l@{}}1856 \\(+1594) \end{tabular} & manual\&auto
\end{tabular}}
\subcaption{Overview of \textsc{DEplain}, the proposed German simplification corpus on document- and sentence-level. }
\label{table-german-datasets-deplain}
\end{subtable}
\caption{Overview of German simplification corpora. All corpora contain German languages (no dialect specified) except 20Minuten (see $^*$, Swiss German, de-CH), APA-benchmark, APA-LHA, \textsc{DEplain-apa} (see $^\ddagger$, Austrian German, DE-AT).  All source texts from all corpora address a general audience, except simple-path (see $^\dagger$).}
\label{table-german-datasets}
    
\end{table*}

\section{Worse Examples of APA-LHA}
\label{appendix-apa-lha-errors}

In the APA-LHA corpus~\cite{spring-etal-2021-exploring} we found original sentences repeated several times in the training data aligned to multiple different simplified sentences (in \autoref{tab-exp-apa-lha} called ``simplifications''). This format can either comprises different simplifications for the same complex sentence or a split of one complex sentence into several simple sentences.\par 
If we see these simplifications as alternative simplifications, some original-simple pairs seems to be wrongly aligned.  In \autoref{tab-exp-apa-lha}, we show two alignment pairs in which the meaning is heavily changed. In the first example, the original and all simplifications are related to career but at different states, i.e., looking back at the career, starting the career and quitting the career. In the second example, the terms of unemployment and short time are mixed and also the numbers are totally different. 

In row three and four of \autoref{tab-exp-apa-lha}, we provide more examples regarding the unclear format, it is not clear whether pairs with identical complex sentences are alternative simplifications (references) (see row 3) or 1:$m$ alignments (see row 4). 

\begin{table*}
\begin{subtable}[c]{\textwidth}
\resizebox{\textwidth}{!}{
\begin{tabular}{p{3.25cm}|p{4cm}p{4.5cm}l}
\textbf{complex-ids} & \textbf{Error Description} & \textbf{Original} & \textbf{Simplifications} \\\hline\hline
A2\#35;A2\#235& Original and all simplifications are related to ``Karriere'' (career) but with different meaning. & ``Auf seine Karriere blickte er gerne zurück .'',  & \begin{tabular}[c]{@{}l@{}}``Er begann seine Karriere mit 18 Jahren .'',  \\ ``Er beendet jetzt seine Karriere .''\end{tabular}\\
 A2\#2621;A2\#265; A2\#6530;A2\_dev\#85; A2\_dev\#417  & The simplifications contain a mix of i) ``unemployment'' and ``short time work'', ii) different digits, and iii) misalignment across documents. & ``Derzeit sind damit aber noch immer über 123.000 Personen mehr arbeitslos als vor der Coronakrise .'' & \begin{tabular}[c]{@{}l@{}}``Derzeit sind in Österreich auch noch 1,3 Millionen Menschen in Kurz-Arbeit .'', \\ ``Die Kurz-Arbeit gibt es , damit nicht noch mehr Menschen ihre Arbeit verlieren .'', \\ ``Dort gibt es jetzt 5.000 Arbeitslose weniger als vor einer Woche .'', \\ ``Vor einer Woche waren noch 9.000 Menschen mehr arbeitslos .'', \\ ``Ihr Geld bekommen sie aber nicht mehr von den Firmen , sondern vom Staat .''\end{tabular} \\ \hline
B1\#3392;B1\#6763 & 

The simplifications can be interpreted as a split of the original sentence into two sentences (one 1:m simplification). 
& ``Laut Polizei dürfte das Kind nach dem im Garten für den Vierbeiner abgelegten Futter gegriffen haben , als es zu der Attacke kam .'' & \begin{tabular}[c]{@{}l@{}}``Laut Polizei griff der Bub im Garten nach dem Hundefutter .'',  \\ ``Dabei kam es zu der Attacke .'' \end{tabular} \\
A2\#246; A2\#5966 & 
The simplifications can be interpreted as alternative simplifications (two 1:1 simplifications). 
& ``Menschen können Hunde und Katzen mit Coronavirus anstecken'' & \begin{tabular}[c]{@{}l@{}}``Menschen können Hunde und Katzen mit dem Corona-Virus anstecken .'',\\  ``Hunde und Katzen können mit dem Corona-Virus angesteckt werden .' \end{tabular}\\

\end{tabular}}
\subcaption{Original German version.}
\end{subtable}
\newline
\vspace*{0.25 cm}
\newline
\begin{subtable}[c]{\textwidth}
\resizebox{\textwidth}{!}{
\begin{tabular}{p{3.25cm}|p{4cm}p{4.5cm}l}
\textbf{complex-ids} & \textbf{Error Description} & \textbf{Original} & \textbf{Simplifications} \\\hline\hline
A2\#35;A2\#2352 & Original and all simplifications are related to ``Karriere'' (career) but with different meaning. & ``He looks back on his career with pleasure .'' & \begin{tabular}[c]{@{}l@{}}``He began his career at the age of 18 .'', \\ ``He is now ending his career .''\end{tabular} \\
A2\#2621;A2\#265; A2\#6530;A2\_dev\#85; A2\_dev\#417  & The simplifications contain a mix of i) ``unemployment'' and ``short time work'', ii) different digits, and iii) misalignment across documents. & ``At present , however , this still leaves over 123,000 more people unemployed than before the Corona crisis .'' & \begin{tabular}[c]{@{}l@{}} ``Currently, 1.3 million people in Austria are also still in short-time work .'', \\
``The short-time work exists so that more people do not lose their jobs .'', \\
``There are now 5,000 fewer unemployed there than a week ago .'', \\
``A week ago , 9,000 more people were unemployed .'', \\
``But they no longer get their money from the companies , but from the state.''\end{tabular}\\ \hline
B1\#3392;B1\#6763 & The simplifications can be interpreted as a split of the original sentence into two sentences (one 1:m simplification). & `According to police , the child may have reached for the food placed in the garden for the quadruped , when it came to the attack .'' & \begin{tabular}[c]{@{}l@{}}``According to police , the boy reached for the dog food in the garden .'', \\
In the process, the attack occurred .'' \end{tabular} \\
A2\#246;A2\#5966 & The simplifications can be interpreted as alternative simplifications (two 1:1 simplifications). & ``People can infect dogs and cats with coronavirus'' & \begin{tabular}[c]{@{}l@{}}``Humans can infect dogs and cats with the Corona virus . '', \\
``Dogs and cats can be infected with the Corona virus .''\end{tabular} 

\end{tabular}}
\subcaption{Translated English version.}
\end{subtable}

\caption{Excerpt of worse automatically aligned sentence-level pairs in the training data of APA-LHA~\cite{spring-etal-2021-exploring}.  All examples are contained in the training data. The complex-ids are a concatenation of the name of the dataset (either C2 to A2 or to B1) and the line number. }
\label{tab-exp-apa-lha}
\end{table*}

\section{Examples of Mild and Strong Simplifications}
\label{appendix-mild-strong-examples}
In \autoref{tab-app-example-simp}, some examples of strong and mild simplifications of \textsc{DEplain} are provided including English translations.

\begin{table*}[]
\resizebox{\textwidth}{!}{
\begin{tabular}{l|p{5cm}p{5cm}p{5cm}p{5cm}lllp{2cm}}
\textbf{} & \textbf{original} & \textbf{simplification} & \textbf{original (English)} & \textbf{simplification (English)} & \textbf{domain} & \textbf{OL} & \textbf{SL} & \textbf{source} \\\hline\hline
\textbf{mild} & Innenminister Herbert Kickl gab am Donnerstag bekannt, dass im Jahr 2018 jedes 2. Verbrechen aufgeklärt wurde. & Außerdem wurde im Jahr 2018 jedes 2. Verbrechen aufgeklärt. Das gab Innenminister Herbert Kickl am Donnerstag bekannt. & Interior Minister Herbert Kickl announced Thursday that every 2nd crime was solved in 2018. & In addition, every 2nd crime was solved in 2018. This was announced by Interior Minister Herbert Kickl on Thursday. & news & B1 & A2 & Austria Press Agency \\
\textbf{mild} & Da entstand Helligkeit. & Und es wurde hell. & That's when brightness arose. & And it became bright. & bible & C2 & A1 & Offene Bibel \\\hline
\textbf{strong} & Über dem Tisch, auf dem eine auseinandergepackte Musterkollektion von Tuchwaren ausgebreitet war – Samsa war Reisender – hing das Bild, das er vor kurzem aus einer illustrierten Zeitschrift ausgeschnitten und in einem hübschen, vergoldeten Rahmen untergebracht hatte. & Auf dem Tisch sind noch immer die Stoffe ausgebreitet. Gregor ist von Beruf Vertreter. Seine Aufgabe ist es, Stoffe zu verkaufen. Dafür reist er umher. Gregor sieht sich weiter in seinem Zimmer um. Über dem Tisch hängt immer noch das Bild. Das Bild, das er vor ein paar Tagen aus einer Zeitschrift ausgeschnitten hat. Gregor hat es in einem schönen Rahmen aufgehängt. In einem goldenen Bilder -Rahmen. & Above the table on which was spread an unpacked sample collection of drapery - Samsa was a traveler - hung the picture he had recently cut out of an illustrated magazine and placed in a handsome gilt frame. & On the table, the fabrics are still spread out. Gregor is a salesman by profession. His job is to sell fabrics. For that, he travels around. Gregor continues to look around his room. Above the table still hangs the picture. The picture he cut out of a magazine a few days ago. Gregor has hung it in a beautiful frame. In a golden picture frame. & fiction & C2 & A2 & Spaß am Lesen Verlag \\
\textbf{strong} & Solange keine vollständige Belastung möglich ist, muss eine Thromboseprophylaxe durch die Gabe von niedermolekularem Heparin durch Spritzen erfolgen. & Der Arzt gibt Ihnen alle wichtigen Informationen. Sprechen Sie deshalb mit Ihrem Arzt. Eine Thrombose ist gefährlich. Die Spritzen sind gegen eine Thrombose. Deshalb müssen Sie vielleicht Spritzen bekommen. Dann dürfen Sie nicht mit dem Fuß auftreten. Manchmal müssen Sie den Fuß mehrere Wochen schonen. & As long as complete weight-bearing is not possible, thrombosis prophylaxis must be given by administration of low-molecular-weight heparin by injection. & The doctor will give you all the important information. Therefore, talk to your doctor. Thrombosis is dangerous. The injections are against thrombosis. Therefore, you may have to get injections. Then you must not step with your foot. Sometimes you need to rest the foot for several weeks. & health & C2 & A2 & Wort \& Bild Verlag Konradshöhe GmbH \& Co. KG
\end{tabular}}
\caption{Examples of mild (upper part) and strong simplifications (lower part) in different domains including the CEFR level of the original (OL) and the simplification (SL).}
\label{tab-app-example-simp}
\end{table*}

\section{Inter-Annotator Agreement}
\label{appendix-tab-iaa}
In \autoref{appendix-table-iaa}, we show an overview of the inter-annotator agreement per domain.

\begin{table}[htb]
\resizebox{\columnwidth}{!}{
\begin{tabular}{l|lllll}
\textbf{domain} & \textbf{avg.} & \textbf{std.} & \textbf{interpretation} & \textbf{\# sents} & \textbf{\# docs} \\\hline\hline
bible & 0.7011 & 0.31 & moderate & 6903 & 3 \\
fiction & 0.6131 & 0.39 & moderate & 23289 & 3 \\
health & 0.5147 & 0.28 & weak & 13736 & 6 \\
language learner & 0.9149 & 0.17 & almost perfect & 18493 & 65 \\
news & 0.7497 & 0.28 & moderate & 25224 & 10 \\\hline
all & 0.8505 & 0.23 & strong & 87645 & 87
\end{tabular}}
\caption{Inter-annotator agreement per domain including average, standard deviation, number of sentence combinations (\# sents), and number of documents (\# docs).}
\label{appendix-table-iaa}
\end{table}

\section{Details on \textsc{DEplain-web}}
\label{sec-deplain-web-process}
In this section, we will describe more details on the web harvester and the process of creating the dataset. 

\subsection{Overview of Web Pages}
The web pages in \autoref{table-deplain-doc-overview} were crawled for generating \textsc{DEplain-web}. We selected these pages based on a web research regarding web pages in German plain language (``Einfache Sprache''). We further checked the references of translation offices, e.g., which web pages are simplified by them and if they contain parallel alignments.

\begin{table*}[htb]
\resizebox{\textwidth}{!}{
  \rowcolors{2}{gray!15}{white}
\begin{tabular}{l|p{6.5cm}p{6.5cm}lllp{5.5cm}l}
\textbf{subcorpus} & \textbf{website simple} & \textbf{website complex} & \textbf{simple } & \textbf{complex} & \textbf{domain} & \textbf{description} & \textbf{\# doc.} \\\hline\hline
\textbf{EinfacheBücher} & \url{https://einfachebuecher.de/} & \url{https://www.projekt-gutenberg.org/} & PG & SG/OG & fiction & Books in plain German & 15 \\
\textbf{EinfacheBücherPassanten} & \url{https://www.passanten-verlag.de/} & \url{https://www.projekt-gutenberg.org/ }& PG & SG/OG & fiction & Books in plain German & \RS{4} \\
\textbf{ApothekenUmschau} & \url{https://www.apotheken-umschau.de/einfache-sprache/} $^\ddagger$ & \url{https://www.apotheken-umschau.de/einfache-sprache/} & PG & SG & health & Health magazine in which diseases are explained in plain German & 71 \\
\textbf{BZFE} & \url{https://www.bzfe.de/einfache-sprache/} $^\dagger$ & \url{https://www.bzfe.de} & PG & SG & health & Information of the German Federal Agency for Food on good nutrition & 18 \\
\textbf{Alumniportal} & \url{https://www.alumniportal-deutschland.org/services/sitemap/} $^\dagger$ &\url{ https://www.alumniportal-deutschland.org/services/sitemap/} & PG & PG & language learner & Texts related to Germany and German traditions written for language learners. & 137 \\
\hline
\textbf{Lebenshilfe} & \url{https://www.lebenshilfe-main-taunus.de/inhalt/} & \url{https://www.lebenshilfe-main-taunus.de/inhalt/} & ETR & SG & accessibility &  & 49 \\
\textbf{Bibel} & \url{https://offene-bibel.de/} $^\dagger$$^\ddagger$ & \url{https://offene-bibel.de/} & ETR & SG & bible & Bible texts in easy-to-read German & 221 \\
\textbf{NDR-Märchen} & \url{https://www.ndr.de/fernsehen/barrierefreie\_angebote/leichte\_sprache/Maerchen-in-Leichter-Sprache,maerchenleichtesprache100.html} $^\ddagger$ & \url{https://www.projekt-gutenberg.org/} & ETR & SG/OG & fiction & Fairytales in easy-to-read German & 10  \\
\textbf{EinfachTeilhaben} & \url{https://www.einfach-teilhaben.de/DE/LS/Home/leichtesprache\_node.html} & \url{https://www.einfach-teilhaben.de} & ETR & SG & accessibility &  &  67\\
\textbf{StadtHamburg} & \url{https://www.hamburg.de/hamburg-barrierefrei/leichte-sprache/} & \url{https://www.hamburg.de} & ETR & SG & public authority & Information of and regarding the German city Hamburg &  79\\
\textbf{StadtKöln} & \url{https://www.stadt-koeln.de/leben-in-koeln/soziales/informationen-leichter-sprache} & \url{https://www.stadt-koeln.de} & ETR & SG & public authority & Information of and regarding the German city Cologne & 85 \\ 
\end{tabular}}
\caption{This table summarizes the web pages (including metadata) which can be extracted with the web crawler. The line separates the documents in plain German from those in easy-to-read German. \textit{simple} correspond to the language level of the simplified documents, and \textit{complex} of the complex documents, where PG=plain German, ETR=easy-to-read German, SG=standard German, OG=old German. The documents marked with  $\dagger$ are openly licensed and therefore part of \textsc{DEplain-web} (row 2 and row 3 in \autoref{deplain-corpora}). All other documents are part of \textsc{DEplain-web} (row 4 in \autoref{deplain-corpora}). The data provider of the documents marked with $^\ddagger$ explicitely state that their documents are professionally simplified and reviewed by the target group.}
\label{table-deplain-doc-overview}
\end{table*}

\subsection{Document Alignment}
The documents are aligned with three strategies in the following order: i) automatic alignment by the reference to the simple document within the complex documents, ii) automatically matching the titles of the documents on the website, and iii) aligning the documents manually. 
All the books in the fiction domain were manually aligned on the document level as the complex data is provided on another web page (i.e. Projekt Gutenberg\footnote{\url{https://www.projekt-gutenberg.org/}}) than the simplified data (i.e., Spaß am Lesen Verlag\footnote{\url{https://einfachebuecher.de/}}, Passanten Verlag\footnote{\url{https://www.passanten-verlag.de/}}, or NDR\footnote{\url{https://www.ndr.de/fernsehen/barrierefreie_angebote/leichte_sprache/Maerchen-in-Leichter-Sprache,maerchenleichtesprache100.html}}). For the simple books, only a preview was available, therefore we only added the first section of the complex book. If the simplified book summarizes parts of more than the first chapter, the documents might be not comparable. We haven't checked that manually.\par
In addition, to download and align the HTML files, the web crawler also extracts some metadata, the plain text of the documents, and the plain text including paragraph endings.

To further align the documents on paragraph or sentence-level the crawler can be integrated into existing alignment tools, e.g., TS-anno~\cite{stodden-kallmeyer-2022-ts}. \par

\subsection{Technical Details.}
We used Python 3 and the Python Package Beautiful Soup for the HTML data and pymupdf for the PDF data. The code of the web crawler is freely available under the CC-BY-4.0 license and can be accessed via \url{https://github.com/rstodden/DEPlain}.

Because the texts are on editable websites the content may change. Therefore we note the date when we crawled the data, the data can then be extracted using a web archieve, e.g., \url{https://archive.org/web/}. It might be possible that the data is updated or that some sources are not available anymore. 

For each source, we aimed at downloading the complete relevant content of the websites. We removed parts such as navigation, advertisement, contact data, and other unnecessary stuff.

\section{DEplain Alignment Statistics}
\label{appendix-deplain-statistics}
In this section, we show statistics of \textsc{DEplain} regarding the alignment of the manual aligned documents. \autoref{appendex-table-deplain-alignment-manual} summarize numbers of \textsc{DEplain-apa} and \textsc{DEplain-web} with respect to $n:m$ alignments, where $n$ and $m$ are $>$ 0, including rephrasing, splitting, merging, and fusion. \autoref{appendex-table-deplain-alignment-auto} summarize numbers of \textsc{DEplain-apa} and \textsc{DEplain-web} with respect to $n:m$ alignments, where $n$ and $m$ are $\leq$ 1, including copied sentences from the complex to the simplified document as well as deletions in the complex documents and additions in the simplified documents. These oddment pairs were automatically extracted after the documents were manually aligned.

\begin{table}[htb]
\resizebox{\columnwidth}{!}{
\begin{tabular}{l|lllll}
\textbf{Name} & \textbf{\# pairs} & \begin{tabular}[c]{@{}l@{}}\textbf{1:1} \\ \textbf{(rephrase)}\end{tabular} & \begin{tabular}[c]{@{}l@{}}\textbf{1:n} \\ \textbf{(split)}\end{tabular} & \begin{tabular}[c]{@{}l@{}}\textbf{n:1} \\ \textbf{(merge)}\end{tabular} & \begin{tabular}[c]{@{}l@{}}\textbf{n:m} \\ \textbf{(fusion)}\end{tabular} \\\hline
\textbf{\textsc{DEplain-apa}} & 13122 & 9912 & 2360 & 382 & 468 \\
\textbf{\textsc{DEplain-web}}  & 1846 & 863 & 796 & 77 & 110
\end{tabular}}
\caption{Statistics on $n:m$ alignments on manual aligned documents, where $n$ and $m$ are $>$ 0.}
\label{appendex-table-deplain-alignment-manual}
\end{table}

\begin{table}[htb]
\resizebox{\columnwidth}{!}{
\begin{tabular}{l|llll}
\textbf{Name} & \textbf{\# pairs} & \begin{tabular}[c]{@{}l@{}}\textbf{1:1} \\ \textbf{(identical)}\end{tabular} & \begin{tabular}[c]{@{}l@{}}\textbf{0:1} \\ \textbf{(addition)}\end{tabular} & \begin{tabular}[c]{@{}l@{}}\textbf{1:0} \\ \textbf{(deletion)}\end{tabular}   \\\hline
\textbf{\textsc{DEplain-apa}} & 12353 & 2712 & 3964 & 5677   \\
\textbf{\textsc{DEplain-web}} & 5482 & 887 & 1572 & 3050 
\end{tabular}}
\caption{Statistics of additional $n:m$ aligned pairs on manual aligned documents, where $n$ and $m$ are $\leq$ 1. }
\label{appendex-table-deplain-alignment-auto}

\end{table}

\section{Manual Evaluation Rating Aspects}
\label{appendix-rating_en}
In this section, we summarize the aspects used for manual evaluation as well as the accompanied statements. The statements are translated to English, they were shown to the annotators in German. All of the aspects were rated on a 5-point Likert scale, either from -2 to +2 or 1 to 5. An overview of the aspects is shown in \autoref{tab:rating_en}.

\begin{table}[htb!]
\resizebox{\columnwidth}{!}{
\begin{tabular}{p{2.5cm}|p{8.75cm}}

\textbf{item} & \textbf{Statement} \\ \hline
\textbf{Grammaticality} & The simplified sentence is fluent, and there are no grammatical errors.\\ 
\textbf{\begin{tabular}[c]{@{}l@{}}Grammaticality \\ (original)\end{tabular}} & The original sentence is fluent, there are no grammatical errors.\\ 
\textbf{\begin{tabular}[c]{@{}l@{}}Simplicity \\ (simple)\end{tabular}} & The simplified sentence is easy to understand. \\ 
\textbf{\begin{tabular}[c]{@{}l@{}}Simplicity \\ (original)\end{tabular}} & The original sentence is easy to understand. \\ 
\textbf{\begin{tabular}[c]{@{}l@{}}Coherence\\ (simple)\end{tabular}} & The simplified sentence is understandable without reading the whole paragraph.\\ 
\textbf{\begin{tabular}[c]{@{}l@{}}Coherence\\ (original)\end{tabular}} & The original sentence is understandable without reading the whole paragraph.\\ 

\textbf{Meaning Preservation} & The simplified sentence adequately expresses the meaning of the original sentence, perhaps omitting the least important information. \\ 
\textbf{\begin{tabular}[c]{@{}l@{}}Overall \\ Simplicity\end{tabular}} & The simplified sentence is easier to understand than the original sentence. \\ 
\textbf{\begin{tabular}[c]{@{}l@{}}Structural \\ Simplicity\end{tabular}} & The structure of the simplified sentence is easier to understand than the structure of the original sentence. \\ 
\textbf{\begin{tabular}[c]{@{}l@{}}Lexical \\ Simplicity\end{tabular}} & The words of the simplified sentence are easier to understand than the words of the original sentence.\\ 
\end{tabular}}
\caption{Statements of the manual evaluation aspects.}
\label{tab:rating_en}
\end{table}

\section{Description of Adaptations of Alignment Methods}
\label{appendix-alignment-methods}
In this section, we are describing the alignment methods and the adaptations we made. 

\paragraph{LHA} \cite{nikolov-hahnloser-2019-large}
is an unsupervised method that finds $1$:$1$ sentence alignments in monolingual 
parallel corpora where documents don't need to be aligned beforehand. It works with a hierarchical strategy by aligning documents on the first level and then aligning sentences within these documents. This method was recommended by \citet{ebling-etal-2022-automatic} for aligning German parallel documents.
\newpage

Our adaptation of this method is comprised of: 
\begin{enumerate*}[label=\roman*)]
    \item disabling the first level of aligning the documents as we already had the true document alignments, and
    \item modifying the language-dependent tools used within the algorithm to fit the German language (e.g., the stopwords list, the tokenizer model, and the word embeddings model). 
    
\end{enumerate*}

\paragraph{Sentence Transformer} \cite{reimers-gurevych-2020-making}
is a simple straightforward method to find $1$:$1$ sentence alignments by computing cosine similarity between embeddings vectors (produced by a sentence transformer model) of sentences on both sides of the monolingual 
parallel corpora, and then picking the most similar pairs and labeling them as \emph{aligned}.
This method is totally dependent on the used sentence transformer model, and the similarity threshold to set.

Our adaptation of this method is comprised of: 
\begin{enumerate*}[label=\roman*)]
    \item testing with new and different sentence transformer models that are either multi-lingual, i.e, LaBSE\footnote{\href{https://huggingface.co/sentence-transformers/LaBSE}{https://huggingface.co/sentence-transformers/LaBSE}}~\cite{feng-etal-2022-language}, or specially designed for German, RoBERTa\footnote{\href{https://huggingface.co/T-Systems-onsite/cross-en-de-roberta-sentence-transformer}{https://huggingface.co/T-Systems-onsite/cross-en-de-roberta-sentence-transformer}}~\cite{conneau-etal-2020-unsupervised}, and
    \item testing with different threshold values, i.e, 0.7, 0.75, 0.8, 0.9. We got the best precision score (\textit{precision=0.96}) when we used LaBSE at a threshold value of 0.9. 
\end{enumerate*}

\paragraph{Vecalign} \cite{thompson-koehn-2020-exploiting}
is a bilingual sentence alignment method that was designed to align sentences in documents of different languages, however, it was also tested in other works on monolingual 
parallel corpora (e.g., \citet{spring-etal-2022-ensembling}). It has two main advantages, it can produce $n$:$m$ alignments, and it can work with more than 200 languages (as it uses the LASER\footnote{\href{https://github.com/facebookresearch/LASER}{https://github.com/facebookresearch/LASER}}~\cite{artetxe-schwenk-2019-massively} sentence representation model in the background; which is multilingual). We used this model only for sentence alignment and not document alignment.

\paragraph{BertAlign} \cite{10.1093/llc/fqac089}
is a attempt to allow sentence-transformer-based methods to produce $n$:$m$ alignments. It was tested on Chinese-English parallel 
corpora and showed promising results.
Our adaptation of this method was only by using a dedicated German sentence transformer model in the algorithm procedure. Following its outperforming results in the sentence transformers experiment, we used the LaBSE sentence transformer model in this experiment.

\paragraph{MASSAlign}\cite{paetzold-etal-2017-massalign}
is a Python package which includes an easy-to-use alignment method on the paragraph- and sentence-level by \citet{DBLP:journals/corr/PaetzoldS16}.  The method uses a vicinity-driven approach with a similarity matrix based on a TF-IDF model. It is capable of $1$:$1$, $1$:$m$, and $n$:$1$ alignments. Our adaptation to this model are:
\begin{enumerate*}[label=\roman*)]
    \item updating from Python 2 to Python 3
    \item making it more language-independent by flexibly adding a stop word list in the required language.
\end{enumerate*} We don't use the updated version of MASSAlign with Doc2Vec by \citet{paun-2021-parallel} as we only align on the sentence level. In the first experiment, we found out that paragraph alignment is also required for this algorithm.

\paragraph{CATS}~\cite{stajner-etal-2018-cats}
is an alignment method that can also align paragraphs and sentences. CATS align each original sentence with the closest simple sentence by calculating the similarity of all of them based on n-grams (option: C3G) or word vectors (option: CWASA and WAVG). In our experiments CATS only aligned pairs of type $1$:$1$. Officially the code was published in Java\footnote{\url{https://github.com/neosyon/SimpTextAlign}}, for better integrity with the other alignment methods, we used the existing Python version of it\footnote{\url{https://github.com/kostrzmar/SimpTextAlignPython}}. We did experiments with all three options, for the word vectors we used the German embeddings of fasttext\footnote{\url{https://fasttext.cc/docs/en/crawl-vectors.html}}~\cite{athiwaratkun-etal-2018-probabilistic}. 
We just report the best result which was achieved with C3G.

\section{Train, Dev, Test Split for Simplification}
\label{appendix-train-dev-test}
\autoref{table-data-split} provides the size of train, development and test set of \textsc{DEplain}.

\begin{table}[htb]
\resizebox{\columnwidth}{!}{
\begin{tabular}{l|lll|lll}
 &  \multicolumn{3}{c|}{\textbf{document-level}}  &  \multicolumn{3}{c}{\textbf{sentence-level}}   \\
 & \textbf{\textsc{web}} & \textbf{\textsc{apa} }& \textbf{\textsc{apa+web}}&\textbf{\textsc{web}} & \textbf{\textsc{apa}} & \textbf{\textsc{apa+web}}\\\hline\hline
train & 481 & 387 & 868 & 1281 & 10660 & \begin{tabular}[c]{@{}l@{}} 11941  \\ (10660+1281) \end{tabular} \\
dev & 122 & 48 & 170 & 313 & 1231 & \begin{tabular}[c]{@{}l@{}} 1544 \\ (1231+313) \end{tabular}  \\
test & 147 & 48 & - & 1846 & 1231& - 
\end{tabular}}
\caption{Overview of train/dev/test split.}
\label{table-data-split}
\end{table}

\section{Further Results for Simplification.}
\label{appendix-tables-results-sentence-level}
We present results of our models trained on \textsc{DEplain} on existing test sets for German text simplification. In \autoref{app:result-doc-level}, results are shown regarding document simplification and, in \autoref{app-result-sent-level}, regarding sentence simplification.

\subsection{Results on Document Simplification.}
\label{app:result-doc-level}
\autoref{tab-result-20min} shows results of our document-level TS experiments trained on different parts of DEplain using long-mBART with vocabulary reduced to 35k tokens. \emph{APA} correspond to \textsc{DEplain-apa} and \emph{web} to \textsc{DEplain-web}. For a better comparison, we also add the results of a baseline model (last part) and a comparable model reported in \citet{rios-etal-2021-new} (first part, numbers are copied from them).

\begin{table}[htb]
\resizebox{\columnwidth}{!}{

\begin{tabular}{l|llllll}
\textbf{train data} & \multicolumn{1}{c}{\textbf{n}} & \textbf{SARI} & \textbf{BLEU} & \textbf{BS-P} & \textbf{FRE} \\\hline\hline
20min & 18305 & \textbf{33.29} & \textbf{6.29} &   &  \\\hline
DEplain-APA & 387 & 22.805 & 1.706 & 0.03 & 63.9 \\
DEplain-web & 481 & 27.113 & 1.81  & 0.007 & 63.5 \\
DEplain-APA+web & 868 & 24.265 & 1.804 & 0.029 & 64 \\\hline
src2src &  & 1.953 & 2.051 & 0.029 & 54.45
\end{tabular}
}
\caption{Results on Document Simplification Testing on 20min with long-mBART }
\label{tab-result-20min}

\end{table}

\subsection{Results on Sentence Simplification}
\label{app-result-sent-level}
In this section, we present results on existing test sets, i.e., \emph{ZEST}~\cite{mallinson-etal-2020-zero} (see \autoref{appendix-table-results-zest}),  \emph{APA-LHA C2-A2}~\cite{spring-etal-2021-exploring} (see \autoref{table-results-apa-lha-a2}), \emph{APA-LHA C2-B1}~\cite{spring-etal-2021-exploring} (see \autoref{table-results-apa-lha-b1}), and \emph{TCDE19}~\cite{naderi-etal-2019-subjective} (see \autoref{table-results-tcde}).

\begin{table}[htb]
\resizebox{\columnwidth}{!}{
\begin{tabular}{l|lllll}
& \textbf{SARI} & \textbf{BLEU} & \textbf{BS-P} & \textbf{FRE} \\\hline\hline
ZEST & 39.09 & 56.68 & - & - \\
U-NMT & 35.22 & 52.02  & - &-\\
U-SIMP & 40.0	& 61.1 & - &-\\\hline
mBART-APA & \textbf{45.81} &\textbf{56.802} & 0.769 & 67.282 \\
mBART-APA+web &  44.913 & 54.718 & 0.778  & 66.588\\\hline

src2src & 26.812	&67.116	& 0.856 &61.5
\end{tabular}}
\caption{Results on the test set of \emph{ZEST}.
}
\label{appendix-table-results-zest}
\end{table}

\begin{table}[h!tb]
\resizebox{\columnwidth}{!}{
\begin{tabular}{l|lllll}
 & \textbf{SARI} & \textbf{BLEU} & \textbf{BS-P} & \textbf{FRE} \\\hline\hline
Sockeye & \textbf{42.04} & \textbf{15.2}  & - & - \\\hline
mBART-APA & 27.987 & 5.294	&0.232  &57.865\\

mBART-APA+web & 28.468 & 5.464 & 0.236& 56.969\\\hline

src2src & 4.092	&3.635	& 0.184 &44.9

\end{tabular}}
\caption{Results 
on the test set of \emph{APA-LHA C2-A2}.
}
\label{table-results-apa-lha-a2}
\end{table}

\begin{table}[h!tb]
\resizebox{\columnwidth}{!}{
\begin{tabular}{l|lllll}
 & \textbf{SARI} & \textbf{BLEU} & \textbf{BS-P} & \textbf{FRE} \\\hline\hline
Sockeye & \textbf{40.73	}& \textbf{12.3 } & - &- \\\hline
mBART-APA & 29.086&6.495 &0.272 &57.299 \\
mBART-APA+web & 28.527&6.604 &0.273 &56.848\\\hline
src2src & 5.325	&6.18 & 0.236	&44.9 
 
\end{tabular}}
\caption{Results
on the test set of \emph{APA-LHA C2-B1}.
}
\label{table-results-apa-lha-b1}
\end{table}

In each of the tables, the first part includes the results of other models trained on other data than \textsc{DEplain}, the middle part includes the results of our models trained on \textsc{DEplain}, and the last part is the result of the baseline model. The scores of the other models are extracted from the corresponding papers, we do not calculate them ourselves as the model checkpoints or model predictions are not available. Hence, scores of some metrics are missing if they were not reported, e.g., FRE or BERTScore.  Furthermore, different implementations of the metrics might be used, therefore the scores should be interpreted with caution. 

\begin{table}[h!tbp]\resizebox{\columnwidth}{!}{
\begin{tabular}{l|lllll}
 & \textbf{SARI} & \textbf{BLEU} & \textbf{BS-P} & \textbf{FRE} \\\hline\hline
ZEST & \textbf{41.12} & \textbf{21.11}   & - & -\\
U-NMT & 35.97 & 11.72  & - &-\\
U-SIMP & 37.4 & 15.03  &  -&-\\\hline
mBART-APA & 38.964 &	16.85  & 0.539& 44.85\\
mBART-APA+web & 36.937& 16.321 &0.542 &43.65\\\hline
src2src & 14.999	&27.348	 	&0.546 &28.1

\end{tabular}}
\caption{Results 
on the test set of \emph{TCDE19}. 
The scores of the other models are copied from~\citet{mallinson-etal-2020-zero}.}
\label{table-results-tcde}
\end{table}

\newpage
\clearpage

\end{document}